\documentclass[conference]{IEEEtran}
\IEEEoverridecommandlockouts
\usepackage{cite}
\usepackage{amsmath,amssymb,amsfonts}
\usepackage{algorithmic}
\usepackage{graphicx}
\usepackage{textcomp}
\usepackage{xcolor}
\usepackage{subfig}
\usepackage{tikz}
\usepackage{pgfplots}
\usepackage{multirow}
\pgfplotsset{compat=1.17}
\usepackage[symbol]{footmisc}

\def\BibTeX{{\rm B\kern-.05em{\sc i\kern-.025em b}\kern-.08em
    T\kern-.1667em\lower.7ex\hbox{E}\kern-.125emX}}
\begin{document}

\title{Synthetic Sonar Image Simulation with Various Seabed Conditions for Automatic Target Recognition}

\author{\IEEEauthorblockN{Jane Shin}
\IEEEauthorblockA{\textit{University of Florida}\\
Gainesville, FL, USA\\
jane.shin@ufl.edu}
\and
\IEEEauthorblockN{Shi Chang}
\IEEEauthorblockA{\textit{Cornell University} \\
Ithaca, NY, USA \\
sc2892@cornell.edu}
\and
\IEEEauthorblockN{Matthew J. Bays}
\IEEEauthorblockA{\textit{Naval Surface Warfare Center, Panama City Division} \\
Panama City, FL, USA \\
matthew.j.bays2.civ@us.navy.mil}
\and
\IEEEauthorblockN{Joshua Weaver}
\IEEEauthorblockA{\textit{Naval Surface Warfare Center, Panama City Division} \\
Panama City, FL, USA \\
joshua.n.weaver3.civ@us.navy.mil}
\and
\IEEEauthorblockN{Thomas A. Wettergren}
\IEEEauthorblockA{\textit{Naval Undersea Warfare Center, Newport} \\
Newport, RI, USA \\
thomas.a.wettergren.civ@us.navy.mil}
\and
\IEEEauthorblockN{Silvia Ferrari}
\IEEEauthorblockA{\textit{Cornell University} \\
Ithaca, NY, USA \\
ferrari@cornell.edu}
}

\maketitle

\begin{abstract}
We propose a novel method to generate underwater object imagery that is acoustically compliant with that generated by side-scan sonar using the Unreal Engine. We describe the process to develop, tune, and generate imagery to provide representative images for use in training automated target recognition (ATR) and machine learning algorithms. The methods provide visual approximations for acoustic effects such as back-scatter noise and acoustic shadow, while allowing fast rendering with C++ actor in UE for maximizing the size of potential ATR training datasets. Additionally, we provide analysis of its utility as a replacement for actual sonar imagery or physics-based sonar data.
\end{abstract}

\begin{IEEEkeywords}
Automated Target Recognition, sonar, machine learning, simulation.
\end{IEEEkeywords}

\section{Introduction}
\label{sec:intro}

The limited communication environment within the undersea domain has necessitated the use of autonomy and automated target recognition (ATR) in order to allow unmanned vehicles to make actionable decisions without an operator in-the-loop \cite{board2005autonomous, stilwell2000platoons, sariel2008naval}. The underwater environmental properties make acoustic sensors become the most significant sensing tool for developing autonomous systems as shown in vehicle coordination \cite{BaillieulVehicleFormationAcoustic03} and underwater SLAM \cite{FallonUnderwaterForwardSonarSLAM13}. However, the same inhospitable environment makes the collection of large data sets for use in machine learning algorithms to properly train machine learning-based algorithms difficult. As such, significant interest has been in the generation of acoustically accurate data for use in training autonomous systems operating based on side-scan sonar images \cite{sammelmann1995pc, sammelmannpc2014, hunter2005towards,HuoKLSG20}.

\begin{figure}[t]
\centering
  \includegraphics[angle=0,width=0.48\textwidth]{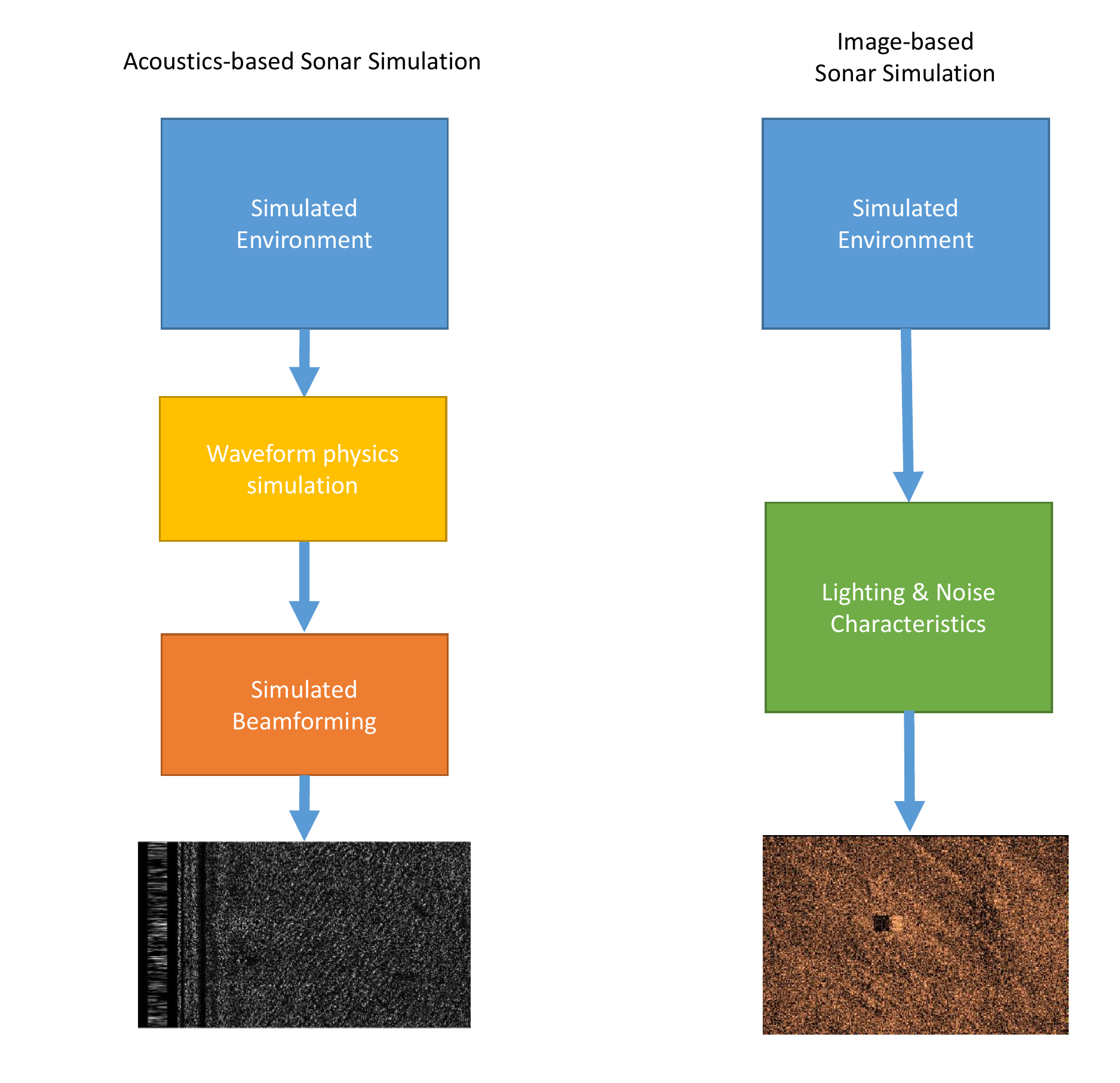}
  \caption{Illustration of process flow for physics-based vs. image-based sonar simulation.}
  \label{fig:illustration}
\end{figure}

One method to generate simulated data is the use of physics-based modeling of the acoustics in order to simulate the sound propagation and raw sonar data collection \cite{choi2021physics, sammelmann1997high}. While this has the benefit of capturing low-level nuances of the sonar data for generation of sonar imagery, the models are typically complex and computationally expensive. An alternative approach is to approximate the imagery that would be generated by sonar beam-forming to directly generate synthetic sonar data \cite{CoirasGPUbasedSim09, woods2020ray, GueriotTubeTracing10}. Figure \ref{fig:illustration} shows a simplified process flow diagram to illustrate the differences between these two approaches.
These existing image-based sonar simulations are easily integrated to computer graphics tool, and this integration allows the simulator to have complicated object geometry more easily. To our best knowledge, however, existing image-based sonar simulations lacks details in modeling the environment, specifically the seabed characteristics. This limitation may become a bottleneck in utilizing simulated images as training data because seabed condition is shown to affect ATR performance \cite{WilliamsFastATR15}.

We propose a novel method to the synthetic sonar generation problem by using the Unreal Engine (UE) to approximate sonar imagery in a large-scale complicated underwater environment and including back-scatter noise and other acoustic effects. We present our approach to properly tuning the UE settings to provide the appropriate shadow effects often found in side-scan sonar, our method to train against the simulated data, and analysis of its performance in computational complexity and transfer learning. One of the main contribution of our simulation tool is to provide multiple seabed conditions, which affect the performance in training ATR or ML algorithms.

Our paper is outlined as follows: Section \ref{sec:RelatedWork} discusses the state-of-the-art sonar simulation used in ATR from both physics-based and image-based sonar simulation tools. We then discuss the overall setup and architecture of our UE-based simulation environment in Section \ref{sec:UnrealEngine}. Finally, we provide analysis of the UE simulation tool in Section \ref{sec:Results} based on the computation time of synthetic sonar image generation and transfer learning performance of an ATR algorithm.

\section{Related work}
\label{sec:RelatedWork}

Early, simple simulation models have been developed to implement sonar imaging patterns (highlights and shadows) of simple object geometry \cite{BellThesisSidescanSonarSimModel95,BellOpticalRayTrackingSonarSim97, BellSimulationAnalysisSyntheticSidescanSonar97}. For real-time simulated sonar image generation, early work focuses on capturing the highlight and shadow patterns of simple object geometry without computing realistic sonar intensity values in the simulated sonar images \cite{GuImageSonarSimForOBjectRecognition13, PailhasRealTimeSideScanSim09}. Sonar image simulators with this simple model enables smooth integration of sonar sensor modules into existing open-source robotics simulators, such as Gazebo using ROS nodes \cite{MahaesGazeboUUVSim16, DeMarcoGazeboSonar2D15}. This feature also allows easier adoption of additional modules for various underwater robotics applications \cite{McconnellOverhead22}.

However, such simplistic tools often fail to properly account for complicated acoustic effects found in the beamformed imagery such as noise and sonar nadir components. In order to obtain more realistic sonar images, high-fidelity physics-based simulators have been developed \cite{choi2021physics}. One of the earliest of these examples is the Shallow Water Acoustics Toolset for a Personal Computer (PC-SWAT) \cite{sammelmann1997high}. This work leverages a high frequency sonar performance prediction model of the shallow water and very shallow water environment to develop the necessary beamforming to then create simulated acoustic images.

More recently, work has been developed to model more complicated object geometry and leverage computer graphic tools that implement the ray-tracing for sonar simulation. This approach allows the simulation to model more complicated scenes \cite{Potokar22icra}. However, the ray-tracing model has limitations in implementing accurate acoustic propagation and relevant acoustic noise. These shortcomings have been sought to be overcome by recent work by Woods \cite{woods2020ray}. However, physics-based sonar image simulators require high computational complexity. Although there has been research on overcoming this high computational complexity using tube-tracing \cite{GueriotTubeTracing10} and GPU-based computation \cite{CoirasGPUbasedSim09}, this limitation can still affect automatic target recognition performance because it is expensive to generate a large training data set.

Moreover, existing sonar imaging simulators do not explicitly consider seabed conditions to the best of our knowledge. Automatic target recognition algorithms are trained by learning the highlight-shadow patterns of the sonar images of an object. However, in real applications, highlight-shadow patterns vary significantly depending on seabed conditions, such as sand ripples, rocky seabed, and muddy seabed. For example, when an object lies in an area with sand ripples, the highlight-shadow patterns can be less visible and more complicated to interpret due to the shadow of sand ripples. Therefore, seabed conditions must be considered in the sonar image simulations to incorporate environmental factors that affect underwater perception performance in a real-world setting.

Our work seeks to bridge the gap between the speed and utility of image-based sonar simulation tools found in the first generation of attempts to simulate sonar imagery, with the realism and anticipated performance of physics-based simulation tools. Additionally, by leveraging the open-source UE toolset as a baseline, we allow for the ocean community to augment and improve our baseline toolset in the future. 

\section{Unreal Engine Simulation}
\label{sec:UnrealEngine}

We will now provide an overview of our UE simulation framework developed for sonar simulation. We first discuss general environmental setup of the UE environment to approximate seafloor characteristics and geometry. We then turn to the rendering settings we have found to be most favorable for sonar simulation. Following the rendering, we discuss our addition of noise that approximates that found in acoustic effects such as back-scatter that allows the images to appear similar to those found in actual sonar data.

\begin{figure}[t]
    \centering
    \includegraphics[width=\linewidth]{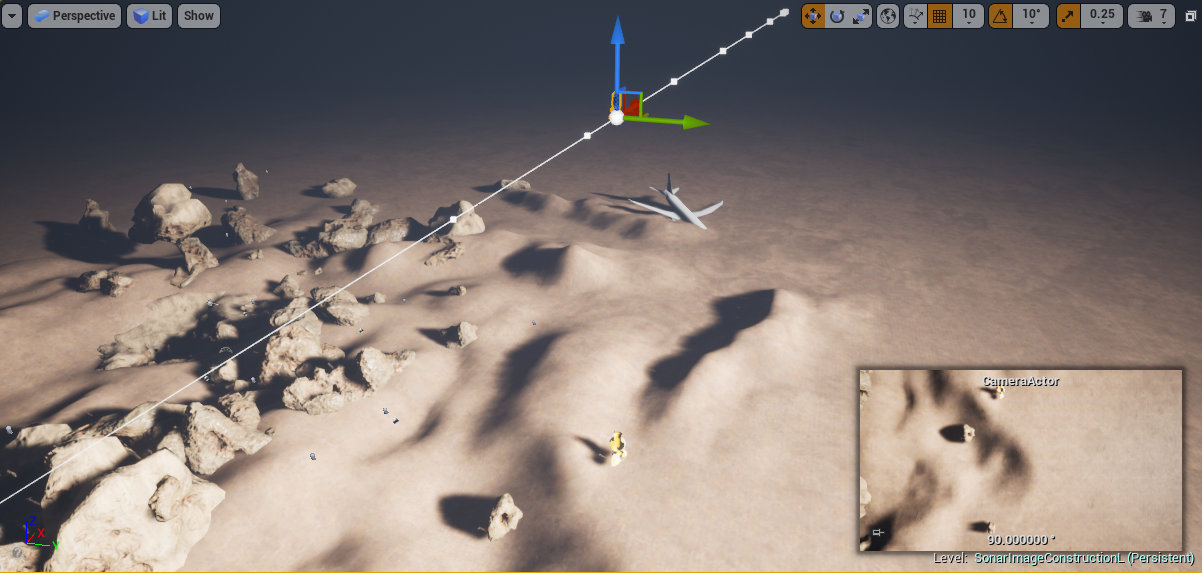}
    \caption{A screenshot of the presented UE simulation tool. The simulation environment includes rocky seabed and objects such as small debris and an airplane. The white solid line with spherical points is the path of the camera actor, whose pose is denoted using blue-green-red axis. The image that the camera actor renders is shown in the lower-right side of the Viewport screenshot.}
    \label{fig:uescreenshot}
\end{figure}

\subsection{UE Simulation Environment Setup}

The UE sonar image simulation tool consists of three major components: a map, objects, and camera and lights setting (Figure \ref{fig:uescreenshot}). A map represents the large underwater environment or the scene that includes underwater objects. The objects inside our simulation tool refer to the targets of interest that need to be detected and identified by ATR or ML algorithms. A camera actor is placed inside the map considering the sonar field-of-view (FOV) to render synthetic sonar images. The lights are set up such that the shadow direction aligns with the camera frame in order to simulate acoustic shadows. The intensity of lighting is manually tuned such that the overall intensity distribution inside the rendered synthetic image can capture acoustic highlight-shadow patterns depending on the texture and materials of objects and the seabed. Thus, lights and camera actor are set up dependently.

\begin{figure}[htbp]
    \centering
    \subfloat[\label{fig:ripple}]{%
        \includegraphics[width=\linewidth]{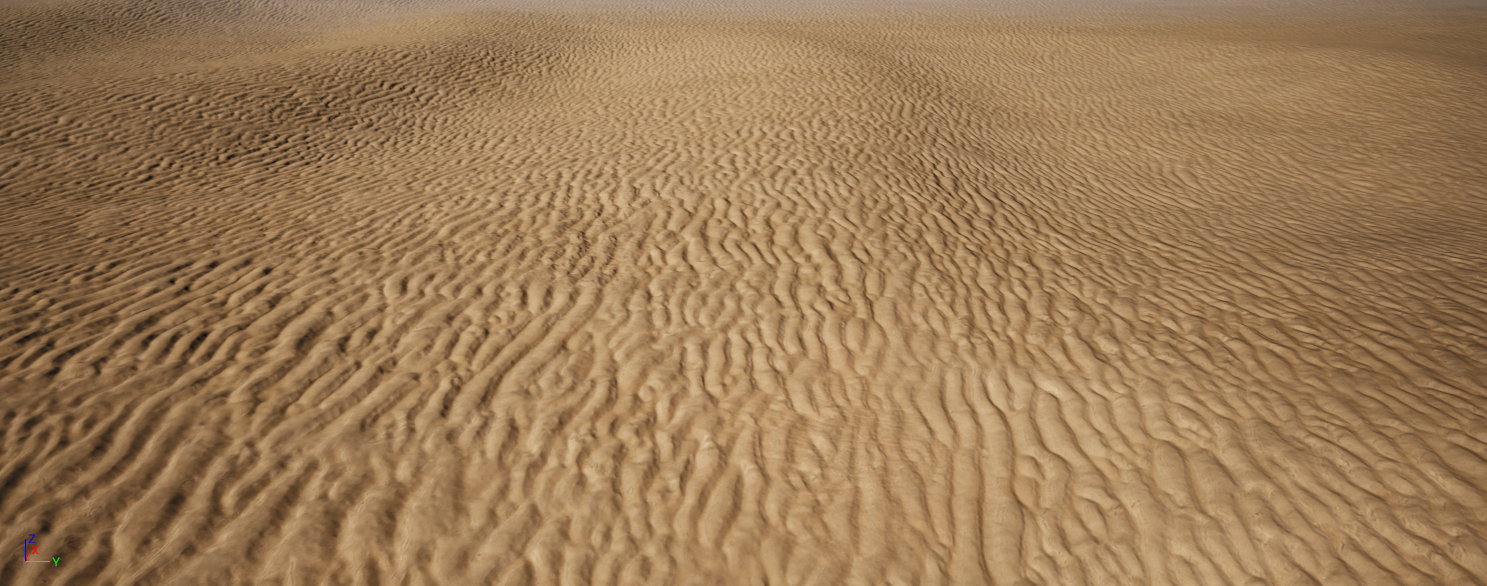}}
    \\
    \subfloat[\label{fig:mud}]{%
        \includegraphics[width=\linewidth]{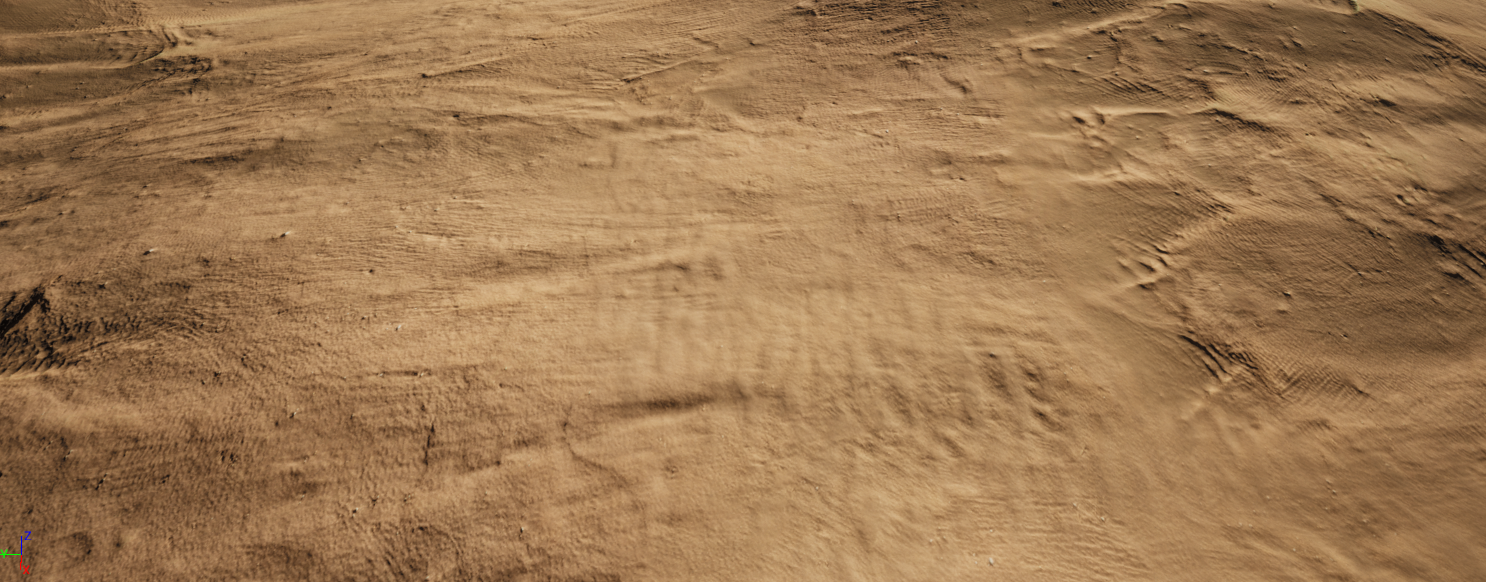}}
    \\
    \subfloat[\label{fig:rock}]{%
        \includegraphics[width=\linewidth]{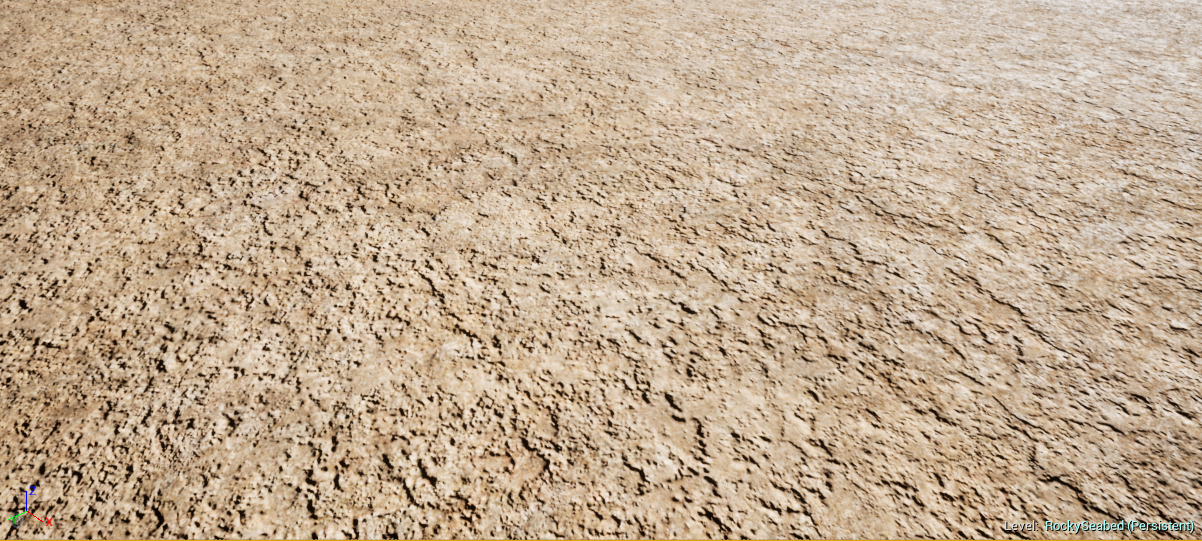}}
    \caption{UE Viewport screenshots of different seafloor conditions in our simulator: (a) sand ripples, (b) muddy and (c) rocky seabeds.}
  \label{fig:ue-env} 
\end{figure}

Our simulation tool allows for three seabed types, sand ripples, mud, and rocks, for synthetic sonar image generation (Figure \ref{fig:ue-env}). Each seabed texture is created inside the UE editor using the Sculpt mode, which allows users to customize the landscape in large scale. Users can also create a more realistic seabed environment by including various seabed types in one map using the Sculpt mode. Moreover, UE Marketplace provides various environments created by other artists such that users can adopt and customize the landscape using the landscape tools. The seabed maps introduced in this paper are adopted and modified from the Marketplace asset in \cite{marketplace22}. In UE editor, each map with a seabed texture is saved as a level inside a UE Project. For the purpose of automatic and fast synthetic sonar image generation, three different levels were created such that each map includes one seabed type.

\begin{figure}[htbp]
    \centering
    \includegraphics[width=\linewidth]{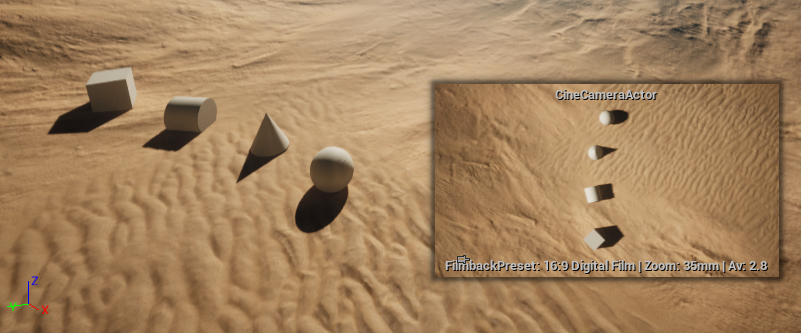}
    \caption{The objects in cube, cylinder, cone, and sphere geometry are placed in a map consisting of different seabed conditions. The image that the camera actor in our simulator renders is shown on the lower-right side.}
    \label{fig:obj}
\end{figure}

Our simulation tool can include various objects either by importing 3D geometry and animation data in various format, such as Filmbox (FBX) and Object (OBJ) file format, or by directly creating an object inside the UE editor. While objects with complicated geometry, such as airplane or shipwreck, can be imported into our simulation, we use the simple objects provided in UE editor for the ATR purpose. These objects can also be customized by setting scales, aspect ratios, and rotations. Another advantage of our simulation tool is that the simulation does not require additional texture file for rendering. The UE editor sets a white color texture on the object by default. We keep this default texture in order to generate highlight patterns in synthetic sonar images. In this paper, we present four different object geometries: block, cone, sphere, cylinder (Figure \ref{fig:obj}).

\begin{figure}[t]
    \centering
    \includegraphics[width=\linewidth]{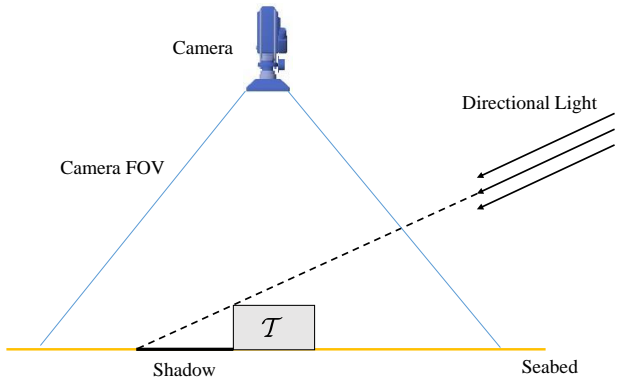}
    \caption{A schematic diagram of the camera and lights setting for synthetic sonar image generator constructed in Unreal Engine simulation. $\mathcal{T}$ denotes the geometry of an object of interest.}
    \label{fig:ImageGenerator}
\end{figure}

Our simulation generates synthetic sonar images by setting the lights and camera actor to mimic the acoustic highlight-shadow patterns. Specifically, the camera is set to face downwards and include the object of interest inside the FOV. While other customized light settings are possible in UE editor, our simulation tool uses optional Sky Light to illuminate the detailed seabed texture and Directional Light to simulate the acoustic shadows. The Directional Lights are set considering the orientation of the vehicle equipped with sonar sensors. Figure \ref{fig:ImageGenerator} shows a schematic diagram of this camera and Directional Light setting. Detailed UE settings can be found in Table \ref{tab:UE_settings}.

\begin{table}[]
\caption{Unreal Engine environment settings. The angle of the Directional Light is measured with respect to the sea floor.}
\begin{tabular}{|c|c|c|}
\hline
                                   & Parameters      & Values                   \\ \hline
\multirow{4}{*}{Camera}            & Projection Type & Perspective              \\ \cline{2-3} 
                                   & Field of View   & $120^{\circ}$                  \\ \cline{2-3} 
                                   & Image Size      & $2048 \times 2048$              \\ \cline{2-3} 
                                   & Capture Source  & Final Color (LDR) in RGB \\ \hline
\multirow{3}{*}{Directional Light} & Angle           & $6^{\circ}$                    \\ \cline{2-3} 
                                   & Intensity       & 30 lux                   \\ \cline{2-3} 
                                   & Light Color     & R 255, G 255, B 255      \\ \hline
Target                             & Material        & Material\_\_85           \\ \hline
\end{tabular}
\label{tab:UE_settings}
\end{table}

\subsection{Rendering in UE using C++ Actor}
Once the UE simulation environment is set up, our sonar simulator generates synthetic images using the rendering engine in UE. UE rendering engine has ray-tracing option, however, our simulator does not use the ray-tracing model because we can still obtain a good quality of synthetic images given the high signal to noise ratio (SNR) of general sonar images. This rendering system allows us to reduce the computation time to render a large amount of images with reasonable quality for the ATR training purpose. When we generate training sonar image datasets, it is important to generate multiple images in different seabed conditions and from different sensor configurations, such as aspect angles and distance from the object \cite{ShinIMVP22}. Therefore, our simulation generates synthetic sonar images by moving and rotating the camera and light setting.

\begin{figure}
    \centering
    \includegraphics[width=\linewidth]{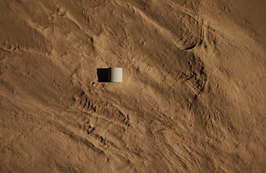}
    \caption{An example of an output image from UE rendering system by setting a camera and a Directional Light as shown in Figure \ref{fig:ImageGenerator}.}
    \label{fig:ex-rgb}
\end{figure}

In order to automate this process of generating training dataset for ATR or ML algorithms, our simulator uses C++ actor in UE. During the depiction of each target, our simulator spawns a 3D model of the specific target on the seabed in a map and under the camera actor using the given target features, target rotations, and pre-constructed 3D prototypes. Our simulator outputs synthetic sonar images by rendering optical images of the target from the camera actor in the presence of directional light. With C++ actor in UE, our simulator can automatically generate a large set of synthetic sonar images for any target from any angles. It is also easy to generate synthetic sonar images for targets in different environmental condition by changing the map representing different seabed conditions. Figure \ref{fig:ex-rgb} show an example of unprocessed image (i.e., rendered image from the UE rendering engine) of a cylindrical target on a muddy seabed.

\subsection{Post-Processing and Acoustic Noise}

\begin{figure}
    \centering
    \includegraphics[width=\linewidth]{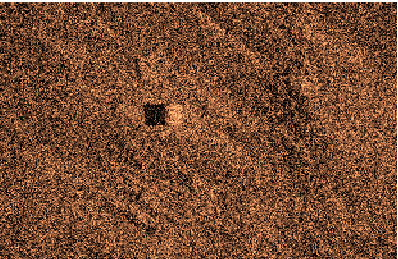}
    \caption{The post-processed synthetic sonar image of Figure \ref{fig:ex-rgb}, which is rendered in UE. In this exsample, Gaussian white noise with mean $0$ and variance of $0.05$ is used.}
    \label{fig:ex-pp}
\end{figure}

The rendered images from the UE simulation are then post-processed to convert the RGB images into the images with intensity values and add acoustic noises. Specifically, our simulation tool includes a separate MATLAB code that inputs the rendered RGB images and then convert that image into grayscale images. If a reference soanr images are provided, a histogram matching step can adjust the intensity of the sythentic to match the reference image. Different acoustic noises are available in our simulation, including Gaussian noise, speckle noise, and Poisson noise. This MATLAB code represents this post-processed image using Copper colormap array in MATLAB for the visualization purposes (Figure \ref{fig:ex-pp}).

By having this post-processing independent from the UE editor, the synthetic images generated from our simulation tool can be easily tuned to different type of sonar sensors by reflecting the sensor and noise properties without changing the whole simulation environment in UE editor. Also, it can be noted that the synthetic sonar images in this paper include one target at the center of each image for the purpose of training ATR algorithms. Note that our simulation tool can also generate more realistic sonar images: for example, synthetic side-scan sonar images can be generated by stitching two images that are obtained in two different Directional Light settings that are in opposite directions and adding the dead-zone between the two images from port and starboard sides.


\section{Results}
\label{sec:Results}
\begin{figure}[th]
    \centering
    \subfloat[\label{fig:cyl1}]{%
        \includegraphics[width=\linewidth]{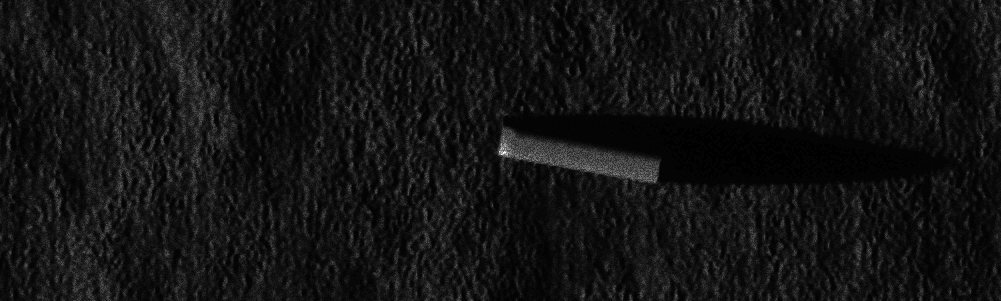}}
    \\
    \subfloat[\label{fig:cyl2}]{%
        \includegraphics[width=\linewidth]{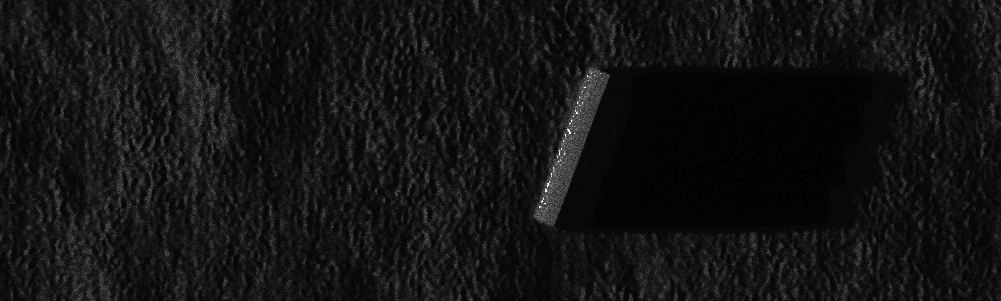}}
    \\
    \subfloat[\label{fig:cube}]{%
        \includegraphics[width=\linewidth]{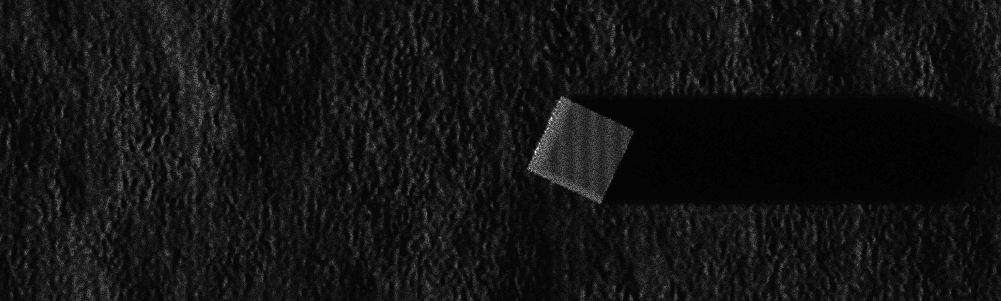}}
        \\
    \subfloat[\label{fig:sphere}]{%
        \includegraphics[width=\linewidth]{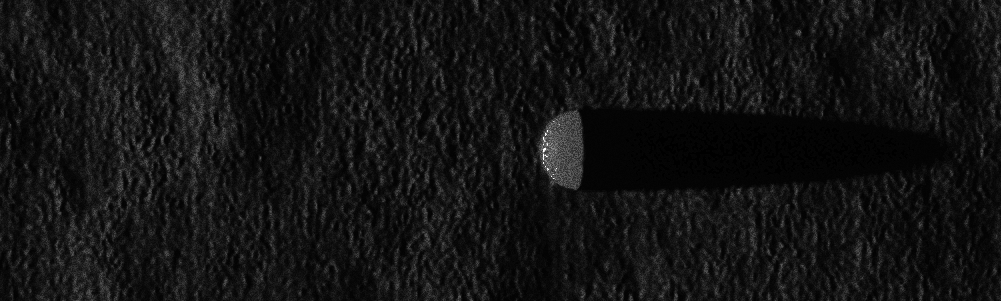}}
    \caption{Synthetic sonar images generated from the presented simulation tool for training an ATR algorithm. (a) and (b) includes a cylinder object in different aspect angles, (c) includes a cube object, and (d) includes a sphere object on a sand ripples.}
  \label{fig:syn-training} 
\end{figure}

\begin{figure}[th]
    \centering
    \subfloat[\label{fig:test-raw}]{%
        \includegraphics[width=\linewidth]{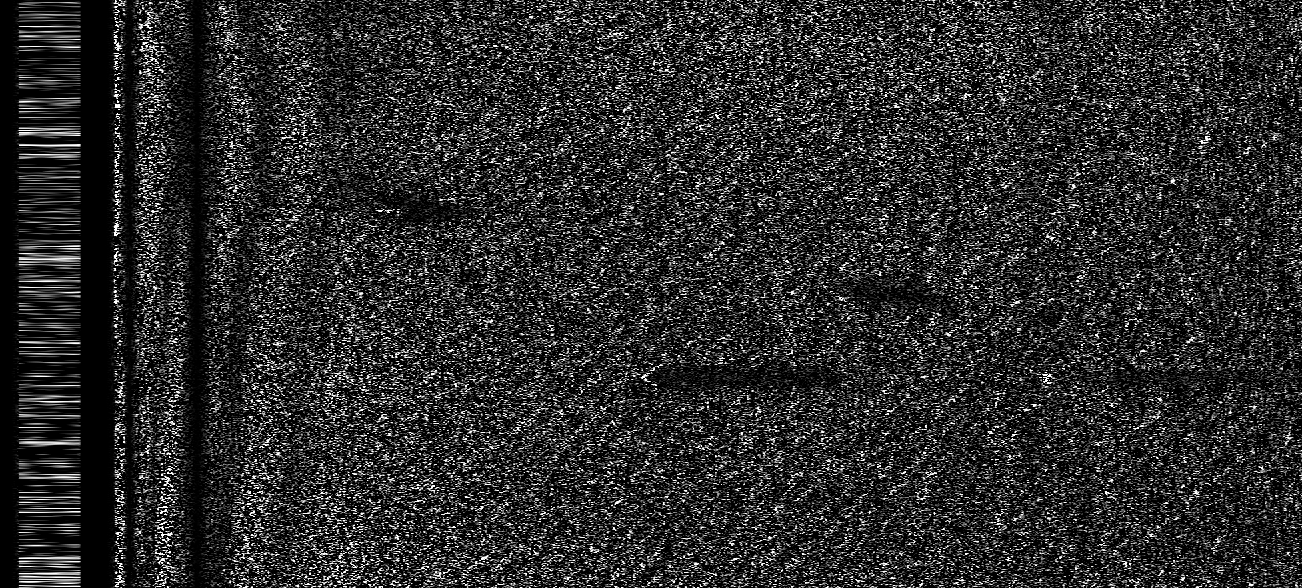}}
    \\
    \subfloat[\label{fig:test-seg1}]{%
        \includegraphics[width=\linewidth]{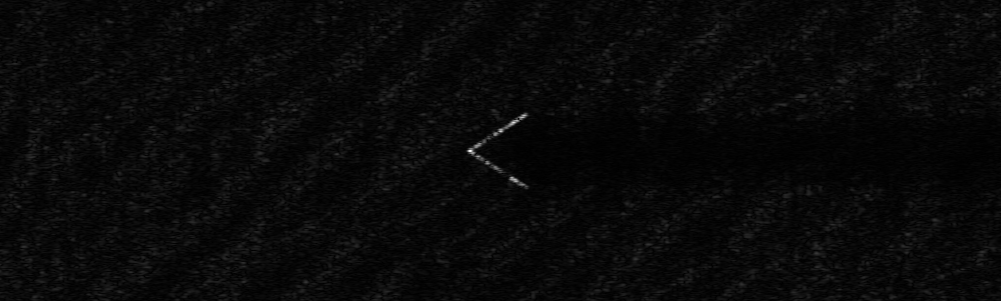}}
    \\
    \subfloat[\label{fig:test-seg2}]{%
        \includegraphics[width=\linewidth]{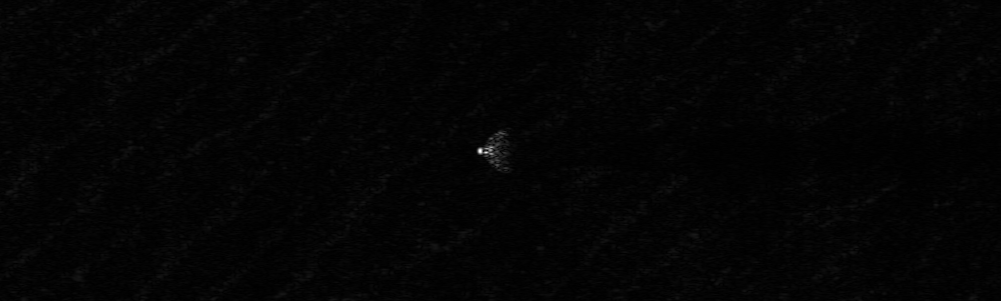}}
    \caption{An image from the test dataset generated from \cite{sammelmann1997high}: (a) a side scan sonar image taken from the port side; (b) a segmented sonar image that includes a block object and sand ripples; (c) a segmented sonar image that includes a sphere object and sand ripples.}
  \label{fig:test-data} 
\end{figure}

The presented sonar image simulation tool is used to generate synthetic sonar images for training ATR algorithms, and the transfer learning performance is evaluated by training an ATR algorithm using the synthetic dataset and testing with sonar images that are generated from a high-fidelity physics-based simulator \cite{sammelmann1997high}. For the evaluation purpose, the simulator is tuned to generate sonar images similar with the test dataset: Figure \ref{fig:syn-training} shows our synthetic sonar images, and Figure \ref{fig:test-data} shows the test images from  \cite{sammelmann1997high}. The generated synthetic sonar images are divided into training and test dataset, and then the training image data are used to train the ATR algorithm proposed in \cite{zhu2017deep}. The trained ATR algorithm is first evaluated with the test dataset consisting of our synthetic sonar images. Then, we evaluate transfer learning performance on a different sonar image dataset. We also evaluate and analyze the computation time in each step for generating sonar images using our simulator.

\subsection{Transfer Learning Performance for ATR}

For the purpose of evaluating transfer learning performance in ATR problem, three target types are used: cylinder, cube, and sphere. The object geometry is chosen based on the test dataset in order to match the profile. Figure \ref{fig:syn-training} shows the synthetic images generated for training the ATR algorithm. For training and testing purpose, total 650 synthetic sonar images are generated for a cylinder object, 600 images are generated for cube object, and 600 images are generated for sphere objects. These images include objects in different aspect angles. After UE rendering system outputs RGB images, we performed an additional histogram matching to the test dataset such that our training data have similar sonar signal intensity. Then, speckle noise with variance 0.1 is added to each synthetic image.

A part of the generated synthetic sonar images are used for training an ATR algorithm. We used the ATR algorithm structure that is introduced in \cite{zhu2017deep} and \cite{chang2018confidence}. This ATR algorithm uses pre-train AlexNet to extract feature vectors from sonar images, and an SVM algorithm is trained using extracted features to perform classification. In order to match the training data with the test dataset, a flat seabed with muddy texture is used in image generation. A total of 80 images are used for training in order to avoid over-fitting, and the remaining synthetic images are used for testing. Specifically, 21 images of a cylinder object, 27 images of a cube object, and 32 images of a sphere object are used for training the ATR algorithm.

In order to test the transfer learning performance, high-fidelity simulated sonar images from \cite{sammelmann1997high} are used as test image data. One of the images is shown in Figure \ref{fig:test-raw}. These raw images are then segmented as shown in Figure \ref{fig:test-seg1} and Figure \ref{fig:test-seg2} for testing the trained ATR algorithm. The test dataset also includes sand ripples and flat seabed conditions. All the objects inside the test dataset are in either cylinder, cube, or sphere geometry.

Figure \ref{fig:confusionmatrix-syn} is the confusion matrix of the ATR algorithm that is trained using our synthetic sonar images and also tested with the remaining synthetic sonar images. Figure \ref{fig:confusionmatrix-act} is the confusion matrix of the ATR algorithm that is trained using our synthetic sonar images and tested with the sonar images from high-fidelity physics-based simulator \cite{sammelmann1997high}. Each category has around 40-60\% accuracy.  Due to the difference between the statistics of the synthetic images and the real sonar images, applying an ATR algorithm trained on the synthetic sonar images directly to the real sonar images may result in a non-satisfactory performance. We have tried to solve this limitation by using more images for training and using more expressive ATR models. Those solution approaches make the ATR algorithm achieve excellent performance when tested on the syntactic images, however, obtain much worse performance on the high-fidelity physical-based sonar images. This limitation is likely due to the ATR algorithm overfitting to the characteristics that only exist on the synthetic images.

However, the ATR algorithm can later be fine-tuned by using a small set of real sonar images. Using the synthetic sonar images for pre-training the classifier can reduce the number of labeled real sonar data required for training an ATR algorithm.   
Moreover, recent work shows that with adversarial training, a machine learning model can be used to refine the simulated camera images \cite{shrivastava2017learning}. We plan to further improve the quality of our synthetic images by adding a generative model based refiner that can reduce the difference between synthetic and real sonar images.

\begin{figure}
    \centering
    \begin{tikzpicture}
        \begin{axis}[
                colormap={bluewhite}{color=(white) rgb255=(90,96,191)},
                xlabel=Predicted,
                ylabel=Actual,
                xticklabels={Cylinder, Cube, Sphere}, 
                xtick={0,...,2}, 
                xtick style={draw=none},
                yticklabels={Cylinder, Cube, Sphere}, 
                ytick={0,...,2}, 
                ytick style={draw=none},
                enlargelimits=false,
                colorbar=false,
                xticklabel style={
                  rotate=0 
                },
                nodes near coords={\pgfmathprintnumber\pgfplotspointmeta},
                nodes near coords style={
                    yshift=-7pt
                },
            ]
            \addplot[
                matrix plot,
                mesh/cols=3, 
                point meta=explicit,draw=gray
            ] table [meta=C] {
                x y C
                0 0 569
                1 0 0
                2 0 50
                
                0 1 0
                1 1 571
                2 1 2
                
                0 2 0
                1 2 0
                2 2 578
            }; 
        \end{axis}
    \end{tikzpicture}
    \caption{Confusion matrix of the ATR algorithm that is trained using synthetic images and tested with exclusive synthetic images.}
    \label{fig:confusionmatrix-syn}
\end{figure}

\begin{figure}
    \centering
    \begin{tikzpicture}
        \begin{axis}[
                colormap={bluewhite}{color=(white) rgb255=(90,96,191)},
                xlabel=Predicted,
                ylabel=Actual,
                xticklabels={Cylinder, Cube, Sphere}, 
                xtick={0,...,2}, 
                xtick style={draw=none},
                yticklabels={Cylinder, Cube, Sphere}, 
                ytick={0,...,2}, 
                ytick style={draw=none},
                enlargelimits=false,
                colorbar=false,
                xticklabel style={
                  rotate=0 
                },
                nodes near coords={\pgfmathprintnumber\pgfplotspointmeta},
                nodes near coords style={
                    yshift=-7pt
                },
            ]
            \addplot[
                matrix plot,
                mesh/cols=3, 
                point meta=explicit,draw=gray
            ] table [meta=C] {
                x y C
                0 0 150
                1 0 129
                2 0 29
                
                0 1 0
                1 1 39
                2 1 49
                
                0 2 0
                1 2 26
                2 2 40
            }; 
        \end{axis}
    \end{tikzpicture}
    \caption{Confusion matrix of the ATR algorithm that is trained using synthetic images and tested with sonar images generated by \cite{sammelmann1997high}.}
    \label{fig:confusionmatrix-act}
\end{figure}

\subsection{Synthetic Sonar Image Generation and Computation Time}
The presented sonar image simulation tool generates synthetic sonar images in UE and post-processes the rendered RGB images in MATLAB to tune the sonar image intensity histogram and add acoustic noises. One advantage of our simulator is that the C++ actor in UE allows us to generate training images in bulk automatically, considering different sensor configuration and aspect angle. In this paper, we run the simulator using Intel Core i7-5960X 3.00GHz 8 Cores 16 Logical cores and Nvidia TITAN Xp Graphics card. The computation time for rendering each image in UE is 1.3747 sec, and the computation time to post-process each image in MATLAB is 0.0691 sec.

\section{Conclusion}

This paper presents a novel underwater sonar imaging simulation tool that generates synthetic sonar images with various seabed conditions and realistic acoustic noise for training ATR and ML algorithms. Various seabed conditions, including sand ripples, mud, and rocks, are implemented in the simulator using Unreal Engine. Sonar-looking highlight-shadow patterns are generated by setting up lights and cameras inside the editor. A large number of images are rendered quickly using the rendering engine and C++ actor in UE. The rendered images are then re-scaled and converted into grayscale images, and artificial acoustic noises are added. The simulator and images generated and used in this paper are available in a public repository\footnote{https://github.com/aprilab-uf/oceans-ue-synthetic-sonar-image}. As a result, simulated synthetic images are used to train underwater automatic target recognition algorithms in this paper. The computation time for image dataset generation and transfer learning performance of the generated sonar images for ATR are evaluated. Future work includes integration of more realistic noise that is dependent on the seabed condition. Moreover, the C++ actor in our simulator can be easily extended and integrated with underwater vehicle simulators. This extension will allow for demonstration of a wide range of robotics applications, for example, adaptive path planning for ATR based on sonar sensor measurements.

\section*{Acknowledgment}
This work was funded by the Office of Naval Research, Code 32. J. Shin and S. Chang contributed equally to this work.

\bibliographystyle{IEEEtran}
\bibliography{oceans22ue}

\end{document}